# Semi-supervised FusedGAN for Conditional Image Generation

Navaneeth Bodla[1]     Gang Hua[2]     Rama Chellappa[1]
Center For Automation Research, University of Maryland, College Park [1]     Microsoft Research [2]
{nbodla,rama}@umiacs.umd.edu    ganghua@microsoft.com

## Abstract

*We present FusedGAN, a deep network for conditional image synthesis with controllable sampling of diverse images. Fidelity, diversity and controllable sampling are the main quality measures of a good image generation model. Most existing models are insufficient in all three aspects. The FusedGAN can perform controllable sampling of diverse images with very high fidelity. We argue that controllability can be achieved by disentangling the generation process into various stages. In contrast to stacked GANs, where multiple stages of GANs are trained separately with full supervision of labeled intermediate images, the FusedGAN has a single stage pipeline with a built-in stacking of GANs. Unlike existing methods, which requires full supervision with paired conditions and images, the FusedGAN can effectively leverage more abundant images without corresponding conditions in training, to produce more diverse samples with high fidelity. We achieve this by fusing two generators: one for unconditional image generation, and the other for conditional image generation, where the two partly share a common latent space thereby disentangling the generation. We demonstrate the efficacy of the FusedGAN in fine grained image generation tasks such as text-to-image, and attribute-to-face generation.*

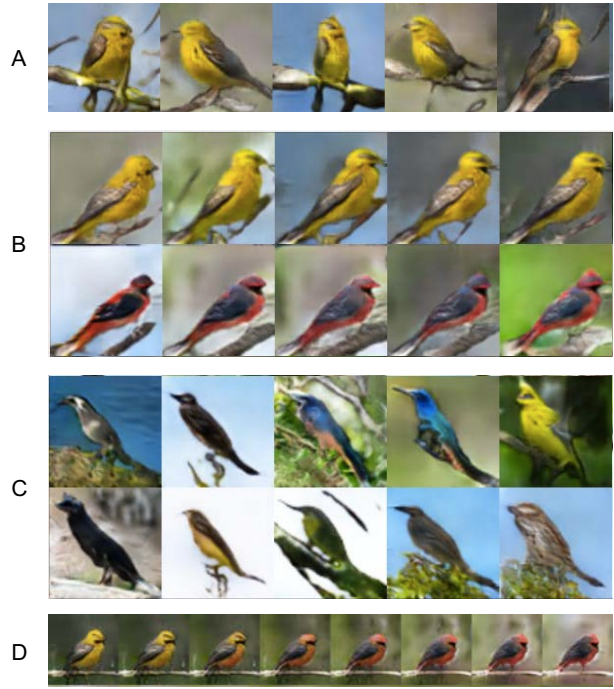

Figure 1: The illustration of sampling with controlled diversity: StackGAN can only generate random images given the corresponding texts as shown in A. In addition to this, our method can generate samples with controlled diversity such as in B, we fix the posture and generate samples with varying details and backgrounds. In C, we fix the posture and generate samples of birds with varying styles as defined by the descriptions. In D, we show examples interpolated between two styles with the same posture.

## 1. Introduction

Recent development of deep generative models has spurred a lot interests in synthesizing realistic images. Generative adversarial networks(GANs) [2] and Variational Autoencoders(VAEs) [6] have been extensively adopted in various applications, such as generating super-resolution images from low resolution images, image inpainting, text-to-image synthesis, attribute to face synthesis, sketch to face synthesis, and style transfer [19, 5, 4], etc.

While synthesizing images by random sampling is interesting, conditional image generation is of more practical value. For example, generating faces given a particular set of attributes as inputs has a lot of practical usage in forensics applications, which makes it easy to make a portrait of a potential suspect. Generating a fine-grained bird image given its description may be of interest in both education and research in biology.

---

*Work done during Navaneeth Bodla's internship at Microsoft Research.

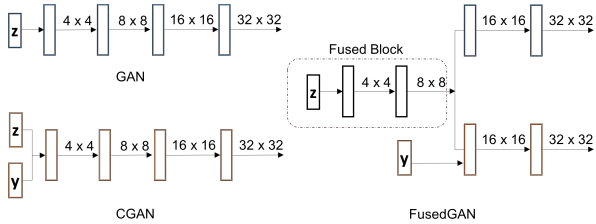

Figure 2: Illustration of FusedGAN by fusing GAN and CGAN for a $32 \times 32$ image synthesis.

CGAN [8] has been widely adopted for synthesizing an image given a condition [18, 17, 1]. A good and effective image generation model needs to possess the following two properties: 1) fidelity and diversity, and 2) controllability in sampling. Controlled sampling refers to the process of sampling images by controlled change of factors such as posture, style, background, and fine-grained details, etc.

By controlling one or more of these factors, diverse images can be generated. For example, one can generate diverse images by keeping a constant background, or generate images with diverse styles by keeping the same posture. Controllability in sampling is directly related to the representation produced from a certain network architecture. We argue that it is equally important to fidelity and diversity, since it can support more practical applications, such as the case we discussed above in generating the portraits of criminal suspects based on describable attributes.

Using text to birds image generation as an example, controllable factors include styles, postures, the amount of fine grained details, and background. Using the StackGAN [18], it is possible to generate birds images with high fidelity, but we have control only over the styles (*i.e.*, text descriptions) in the sampling process. To achieve more controllability in sampling, we need to better disentangle the different factors in the latent space. In attribute2image [17], Yan *et al.* have disentangled the foreground and background generation, and thereby achieving controlled sampling by keeping either one of them fixed and varying the other.

We propose a way to disentangle the structures (which capture the posture and the shape) and the styles (which capture fined-grained appearances of both the foreground and background) to perform image synthesis with high fidelity, diversity and controllability in sampling. Instead of trying to learn a standalone conditional generator, we propose to derive it from an unconditional generator.

We illustrate our approach by a simple thought experiment. Consider the task of painting a bird given a text description, such as "a yellow bird with black wings and a red tail". The most intuitive way of doing this is to first sketch an outline of a bird with a specific posture and shape of the wings, crown, beak and tail. Then, per the description, subsequently paint the wings as black, the body as yellow, and the tail as red. Note that the initial sketch or outline of the bird is independent of the condition, *i.e.*, the text description which defines the style. It is only needed in the later stages of painting the bird.

Motivated by this intuitive process of drawing, and the success of previous stacked deep generative models [16, 18, 20], we propose to disentangle the image generation process such that we learn two cascaded generators. The first unconditional generator produces a structure prior (akin to the initial sketch of the bird) which is independent of the condition, and the second conditional generator further adds style to it and creates an image that matches the condition (check Section 3 for details). In other words, we fuse two generators by cascading them, as shown in Figure 2, where the fused block acts as a structure prior.

By disentangling the generation process, an added advantage of our method is that we can train the unconditional generator using just the images without corresponding conditions. This enables us to conduct semi-supervised data during training. It facilitates in learning a better structure prior (the fused block shown in Figure 2) which in turn contributes to generating better and diverse conditional images.

Our proposed model, referred to as FusedGAN, is unique in the sense that it enjoys the strengths of stacking in a single stage, which can be effectively trained with semi-supervised data. The advantages of our model over existing methods are: 1) it helps in sampling images with controlled diversity. 2) We can use semi-supervised data during our training. This implies that along with usual paired data for conditional image generation such as text to image, we can also leverage images without paired conditions. 2) Unlike FashionGAN [20] and $S^2$GAN [16], we do not require additional intermediate supervision such as segmentation maps or surface normals.

## 2. Related Work

We briefly summarize related works in text-to-image generation, and stacking in deep generative models.

**Text-to-image generation.** Reed *et al* [12] were the first to propose a model called GAN-INT for text to image generation, where they used a conditional GAN to generate images. In their follow-up work GAWWN [11], they leveraged additional supervision in terms of bounding boxes and keypoints to generate birds in a more controlled setup. Zhang *et al.* [18] have extended the idea of conditional GAN to two stage conditional GAN, where two GANs are stacked to progressively generate high resolution images from a low resolution image generated from the first stage. The Stack-GAN is able to produce high resolution $256 \times 256$ images with very good visual quality. Instead of a single-shot image generation, Gregor *et al.* [3] proposed DRAW, which generates images pixel by pixel using a recurrent network.

The key problems, that both GAN-INT [12] and Stack-GAN [18] attempted to address, are diversity and discontinuity of the latent space. Due to the very high dimensionality of the latent space and limited availability of text data, the latent space tends to be highly discontinuous which makes it difficult for the generator to synthesize meaningful images. While GAN-INT proposes a manifold interpolation method during training, StackGAN proposed condition augmentation to sample the text embeddings from a Gaussian distribution.

We further analyze the contribution of condition augmentation in our method, and show that it models the diversity in fine-grained details of the generated birds (check Section 5 for details).

**Stacking.** The core idea behind the proposed FusedGAN model is to disentagle the generation process by stacking. Stacking allows each stage of the generative model to focus on smaller tasks, and disentangling supports more flexible sampling. We briefly summarize previous works addressing disentangling and stacking.

Stacked image generation has shown to be effective in many image synthesis tasks. At a high level, stacked image generation pipelines often have two separate consecutive stages. The first stage generates an intermediate image (such as a segmentation map, or a map of surface normals). Then, the second stage takes the intermediate image as an input to generate a final conditional image. For example, the $S^2$ GAN [16] synthesizes images by first generating the shape structure (*i.e.*, surface normals), and then generates the final image of the scene in the second stage.

StackGAN [18] first generates a low resolution image conditioned on the text embedding $\phi_t$, and subsequently uses it to generate the high resolution image. In fashion-GAN [20], Zhu *et al.* have used the first stage to generate a segmentation map conditioned on the design encoding $\phi_d$ and then used it to generate a new fashion image in the second stage.

We use stacking as a way of learning disentangled representations. Different from these existing work, the stages in our model are implicit. Specifically, in our model, stage 1 performs unconditional image generation and stage 2 performs conditional image generation. However, both stages in our model share a set of high level filters. As a result, the two stages are literally fused into a single stage, which can be trained end to end.

Similar to $S^2$GAN, our model disentangles the style and structure. But different from $S^2$GAN [16], we do not require any additional supervision in terms of surface normals, nor do we require separate training of stages. Similarly FashionGAN [20] and attribute2image [17] both require additional intermediate supervision in the form of segmentation maps, which are not needed in our case.

## 3. The FusedGAN: Formulation

In order to disentangle the generation of the the structure and style, our method comprises of two fused stages. The first stage performs unconditional image generation, and produces a feature map which acts as a structure prior for the second stage. The second stage then generates the final conditional image (*i.e.*, the image that match the style defined by the text description) using this structure prior and the condition as the inputs. It must be noted that there is no explicit hierarchy in stage one and stage two. Both stages can be trained simultaneously using alternative optimization. We use text-to-image synthesis as an example for providing the details of our approach which can be easily extended to other tasks such as attribute-to-face synthesis.

### 3.1. Stage One: Learning a Structure Prior

Our stage one is a GAN where we generate bird images from a random noise vector, and also in the process produces a intermediate representation serving as a structure prior for the second stage. It contains a generator $G_1$ and a discriminator $D_u$, which are pitched against each other in a two player min-max game. In the min-max game, the generator tries to fool the discriminator by generating birds as close to real as possible, whereas the discriminator tries to differentiate between them.

$G_1$ and $D_u$ are both differentiable functions such as deep neural networks and the training is done by optimizing the min-max loss function

$$\min_{G_1} \max_{D_u} V(D_u, G_1) = \mathbb{E}_{x \sim p_{data}}[\log D_u(x)] + \mathbb{E}_{z \sim p_z}[\log(1 - D_u(G_1(z)))]. \quad (1)$$

Since we would like to first generate a structure prior, we split the generator $G_1$ of stage one into two modules: $G_s$ and $G_u$. $G_s$ takes a noise vector $z$ as the input. After a series of convolution and upsampling operations, it generates a structure prior $M_s$. $G_u$ then takes the structure prior as input and again after a series of upsampling and convolutions, generates the final image. Accordingly, $G_1$ in the min-max objective function as presented in Equation 1, is further decomposed to $G_s$ and $G_u$, *i.e.*,

$$M_s = G_s(z), \ G_1(z) = G_u(M_s). \quad (2)$$

where $M_s$ is an intermediate representation. It captures all the required high level information for creating a bird such as the posture and structure. Therefore, it acts as a structure prior that dictates the final shape of the bird. Since the posture and structure information is independent of the style, it could be reused in the second stage to synthesize a bird that matches the description. The advantage of this first stage is that it does not require any paired training data. It can be trained using large datasets containing just the images of the target concept, such as birds for example, which helps in learning an improved structure prior.

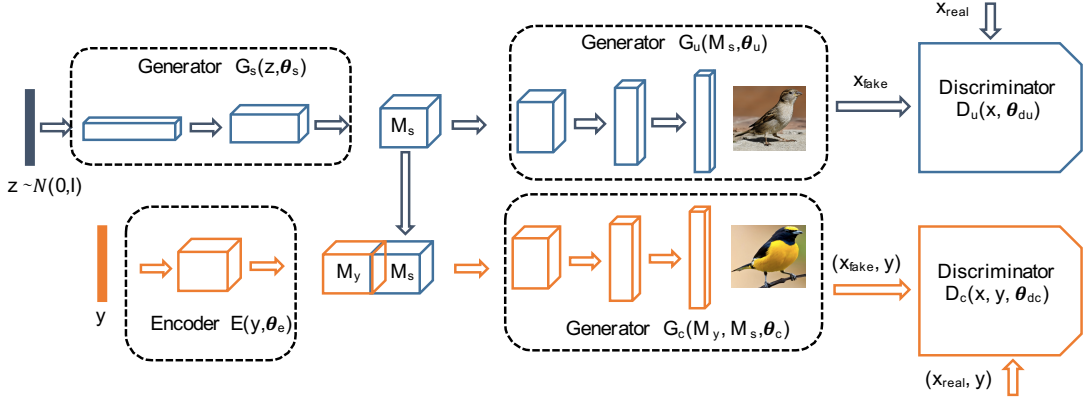

Figure 3: The end to end pipeline of our proposed method. Blue and orange blocks correspond to the unconditional and conditional image generation pipelines respectively.

### 3.2. Stage Two: Stylzing with the Structure Prior

In the second stage, we use a CGAN for generating birds that match the description. Different from the traditional CGAN pipelines, whose input include the condition (*i.e.*, the text description) and the random noise vector, we feed the structure prior $M_s$ from stage one and the text description as inputs to the conditional generator $G_c$. Similar to CGAN, the discriminator $D_c$ of stage two takes an image and condition as inputs to ensure that $G_c$ generates images that match the description.

The $M_s$ acts as a template and provides additional signal to the generator of stage two. This forces the generator to synthesize birds that not only match the description but also preserve the structure information contained in it. Therefore, instead of learning from scratch, $G_c$ builds on top of $M_s$ by adding style to it using the text description. Note that the $M_s$ could also have its own style information from stage one. However, because both the generator and discriminator in stage two takes the text description as inputs, the $G_c$ ensures that the style of the generated image is that of the description and not $M_s$

In this way, the tasks are divided among $G_s$, $G_u$ and $G_c$, where $G_s$ is responsible to learn the overall image structure, and $G_u$ and $G_c$ focus on taking the structure information and generating unconditional and conditional images, respectively. The overall pipeline is shown in figure 3. The conditional GAN is trained by optimizing the following objective function, *i.e.*,

$$\min_{G_c} \max_{D_c} V(D_c, G_c) = \mathbb{E}_{x \sim p_{data}}[\log D_c(x|y)] + \mathbb{E}_{z \sim p_z}[\log(1 - D_c(G_c(G_u(z)|y)))]. \quad (3)$$

**Algorithm 1** The training pipeline of the proposed algorithm.

**Require:** $m$: the batch size; $\theta_s$: initial $G_s$ network parameters; $\theta_u$: initial $G_u$ network parameters; $\theta_e$: initial $E$ network parameters; $\theta_c$: initial $G_c$ network parameters; $\theta_{du}$: initial $D_u$ network parameters; $\theta_{dc}$: initial $D_c$ network parameters.

1: **while** $\theta_c$ has not converged **do**
2:     Sample $x_r \sim P_r$ a batch from the real data;
3:     Sample $z \sim \mathcal{N}(0, I)$ a batch from noise vectors;
4:     $M_s \leftarrow G_s(z)$ ;
5:     $x_{uf} \leftarrow G_u(M_s)$ ;
6:     $\mathcal{L}_{G_u} \leftarrow \log D_u(x_{uf})$;
7:     $\mathcal{L}_{D_u} \leftarrow \log D_u(x_r) + \log(1 - D_u(x_{uf}))$;
8:     $\theta_{du} \xleftarrow{+} -\nabla_{\theta_{du}}(\mathcal{L}_{D_u})$;
9:     $\theta_u \xleftarrow{+} -\nabla_{\theta_u}(\mathcal{L}_{G_u})$;
10:    $\theta_s \xleftarrow{+} -\nabla_{\theta_s}(\mathcal{L}_{G_u})$;
11:    Sample $(x_{cr}, y) \sim P_{cr}$ a batch from the conditional real data;
12:    $M_y \leftarrow E(y)$;
13:    $x_{cf} \leftarrow G_c(M_y, M_c)$;
14:    $\mathcal{L}_{G_c} \leftarrow \log D_c(x_{cf}, y)$;
15:    $\mathcal{L}_{D_c} \leftarrow \log D_c(x_{cr}, y) + \log(1 - D_c(x_{cf}, y))$;
16:    $\theta_y \xleftarrow{+} -\nabla_{\theta_y}(\mathcal{L}_{G_c})$;
17:    $\theta_e \xleftarrow{+} -\nabla_{\theta_e}(\mathcal{L}_{G_c})$;
18:    $\theta_{dc} \xleftarrow{+} -\nabla_{\theta_{dc}}(\mathcal{L}_{D_u})$;
19: **end while**

## 4. The FusedGAN: Learning and Inference

In this section, we provide the details of training our FusedGAN pipeline, as well as the inference procedures. We first present the notation used to describe the training algorithm and then details of the architecture and the inference steps.

**Learning.** Let $\mathbf{z} \in \mathbb{R}^{d \times 1}$ be a noise vector sampled from a normal distribution, *i.e.*, $z \sim \mathcal{N}(0, I)$, where $d$ is the dimen-sionality of the latent space; $G_s(z, \theta_s)$ be the generator that generates the structure prior $M_s \in \mathbb{R}^{s \times s \times k}$; $G_u(M_s, \theta_u)$ be the unconditional image generator that takes the structure prior $M_s$ as input and generates a target image $x_{uf}$; and $D_u(x, \theta_{du})$ be the unconditional image discriminator that takes a real image $x_r$ or a generated image $x_{uf}$ as inputs.

For the conditional image generation pipeline, let $E(\theta_e, y)$ be the text encoder that takes a text embedding $y \in \mathbb{R}^{p \times 1}$ as the input, and produces a tensor $M_y \in \mathbb{R}^{s \times s \times q}$. To achieve this, inspired by the StackGAN [18], condition augmentation is performed to sample latent variables $\hat{c} \in \mathbb{R}^{q \times 1}$ from an independent Gaussian distribution $N(\mu(y), \Sigma(y))$ around the text embedding. The $\hat{c}$ is then spatially repeated to match the spatial dimension of $M_s$ to produce $M_y$.

We denote $G_c(M_y, M_s, \theta_y)$ as the conditional generator that takes $M_y$ and $M_s$ as inputs to generate $x_{cf}$, the conditional image. Similarly, $D_c(x, y, \theta_{dc})$ is the conditional image discriminator which takes a real image $x_{cr}$, or a conditional image $x_{cf}$ along with the condition $y$ as inputs. Both real or generated images are of size $\mathbb{R}^{N \times N \times 3}$.

The standard alternating optimization method is used to train our model. We train the conditional and unconditional pipelines in alternating steps till the model converges as described in algorithm 1. The model parameters are updated by optimizing the combined GAN and CGAN objectives, *i.e.*,

$$
\begin{aligned}
\mathcal{L}_{G_u} &= \log D_u(G_u(z)) \\
\mathcal{L}_{D_u} &= \log D_u(x) \\
\mathcal{L}_{D_c} &= \log D_c(x, y) \\
\mathcal{L}_{G_c} &= \log D_c(G_c(M_y, M_s), y) + \\
&\quad \lambda D_{KL}(N(\mu(y), \Sigma(y)) \| N(0, I))
\end{aligned} \quad (4)
$$

**Architecture.** For the conditional generator $G_c$, $M_s$ and $M_y$ are concatenated and passed through a block containing $3 \times 3$ convolution with stride 1, along with batch normalization and relu to combine the condition and structure prior information. The output is a $8 \times 8 \times 128$ tensor, which is subsequently passed through 4 residual blocks. Each residual block consists of a $3 \times 3$ convolution with output size of 128 and all the convolutions are followed by batch normalization and relu.

The output of the residual blocks is still an $8 \times 8 \times 128$ tensor, which is then progressively upsampled to $16 \times 16$, $32 \times 32$ and $64 \times 64$ to produce the conditional image. All the upsampling operations are followed by $3 \times 3$ convolutions with a stride of 1,along with batch normalization and relu.

The architecture for discriminators $D_u$ and $D_c$ are same as those in StackGAN stage-I, except that for $D_u$ there is no condition concatenation operation. Moreover, the architecture for $G_u$ and $G_s$ combined is same the as stage-I generator of StackGAN. For the text encoder $E$, we use two fully connected layers with batch normalization to produce $\mu(y)$ and $\Sigma(y)$ of 128 dimensions. More details about the architecture are provided in the supplementary material.

The Adam optimizer is used to train our model with a learning rate of 0.0002 and batch size 64. Since we use additional unsupervised data containing just images of the birds, we train the generators for 1000 epochs with a step down after 400 and 800 epochs.

**Inference.** During inference, for generating a conditional image, we first draw a noise sample $z$ from $N(0, I)$, which is passed through $G_s$ to generate the structure prior $M_s$. $M_s$ then takes two paths, one through the generator $G_u$ to produce an unconditional image $x_{uf}$. In the second path, we first send the text input through the encoder $E$, which draws a sample from the Gaussian distribution around the text embedding. The output of $E$ and $M_s$ are concatenated, and passed through $G_c$ to generate the conditional image $x_{cf}$.

Note that in this process, we have two random noise vectors : 1) from $N(0, I)$ and 2) from the distribution of the input text $N(\mu(y), \Sigma(y))$, which are two control factors over the sampling procedure. In other words, in one inference step, we synthesize two images : $x_{cf}$ the conditional image and $x_{uf}$ the unconditional image, a byproduct of our model which helps to analyze and better understand our proposed model and the results.

## 5. Experiments

We present results and analysis of our method in two conditional image generation use cases: 1) text-to-image synthesis using birds as a case study, and 2) attributes-to-image synthesis using faces as a case study. For evaluation of our method we perform both qualitative and quantitative analysis. The qualitative analysis is done by performing user study. For quantitative results, we use the inception score [13].

### 5.1. Text-to-image synthesis

The CUB birds dataset [15] contains 11,788 images. For each image, 10 descriptions and a 1024 dimensional text encodings are provided by Reed *et al.* [10]. The dataset is partitioned into class disjoint train and test splits of 8,855 and 2,933 images, respectively, as mentioned in [12]. Since our approach can handle semi-supervised data, we augment this dataset with the nabirds dataset [14] which contains 48,562 images of the birds without any corresponding text descriptions. We use a total of 57,417 images for our stage one structure prior generation and 8,855 image and text pairs for training the stage two conditional image generator. As a pre-processing step, we crop the images to make sure that object-image size ratio is greater than 0.75 [18].

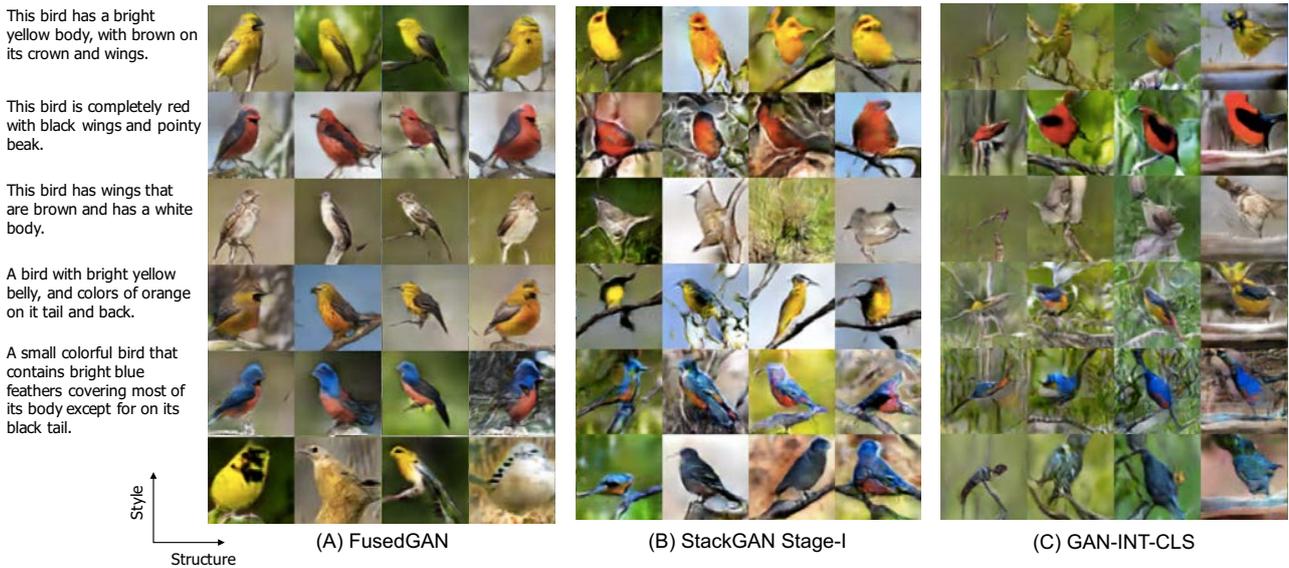

Figure 4: Example birds synthesized from our FusedGAN model, StackGAN stage-I and GAN-INT-CLS. For FusedGAN, the first five rows correspond to the images generated by the respective text descriptions shown on the left and the last row corresponds to the unconditional images generated by our model. For StackGAN and GAN-INT-CLS, the description of the first five rows of the birds match that of FusedGAN but the description for the last row is omitted in the figure.

#### 5.1.1 Results and Analysis

In this section, we present the usefulness of our method in various controlled sampling use cases and compare the performance with two baseline methods: StackGAN stage-I and GAN-INT-CLS. We also provide a detailed ablation analysis on the contributions of various components in our pipeline.

**Fixed posture with varying styles:** Many birds with varying style could have the same posture. We show how to generate them with the FusedGAN. An illustration with visual results are presented in Figure 4 on the left. We also analyze the contribution of the structure prior in the overall conditional image generation process. For this we consider 5 text descriptions $t_i$ where $i = 1, 2, .., 5$ of birds and sample 4 images per description with the same posture in every column as shown in Figure 4 on the left.

In order to control the posture, *i.e.*, to generate birds of various styles (text descriptions) with same posture, we keep the $z$ constant and vary the text descriptions. For example, consider the first column in Figure 4 of FusedGAN. To generate these birds, we sample a $z$ from $N(0, I)$ and pass it through $G_s$ which produces a structure prior $M_s$. We then use the same $M_s$ with 5 of our text description samples to produce the respective conditional images as shown in the first five rows. Notice that they all have the same posture, because the structure prior is the same for them. This demonstrates that the pose and structure information is suc-cessfully captured in $M_s$, and the style information are left to $G_c$.

We further examine the contribution of the structure prior by visualizing the unconditional images, as shown in the last row of Figure 4 for FusedGAN. For the third column, all the birds seem to have a distinct long tail which can also be seen in the unconditional image. Also in the fourth column, we can observe that the unconditional image has a large breast, which is clearly transferred to the yellow, red and orange birds. These results strongly support that $M_s$ is able to successfully capture and transfer significant amount of information about the structure of the bird into the conditional generated bird images of various descriptions.

We further compare the controlled sampling approach with StackGAN and GAN-INT-CLS, as shown in Figure 4. For both methods, we try to control the posture by using the same $z$ as the input to each image in a column, but with varying text descriptions. The GAN-INT-CLS seems to be able to control the posture across all the columns, whereas the StackGAN is not. Although for some results of StackGAN, such as the second column, it seems to have preserved the posture across all styles but for the other columns it does not. For example, in the third column, we can clearly observe that the posture of the last two birds are completely flipped. This indicates that the style and structure are not completely disentangled. In contrast, in results from our FusedGAN, we observe that the structure prior explicitly ensures that the posture is consistently preserved.

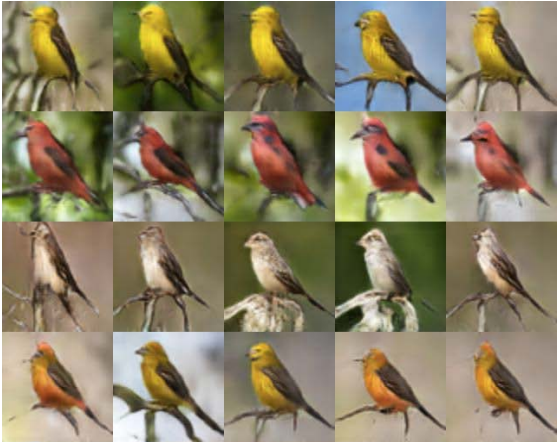

Figure 5: Generated bird images of various styles but varying amount of fine-grained details.

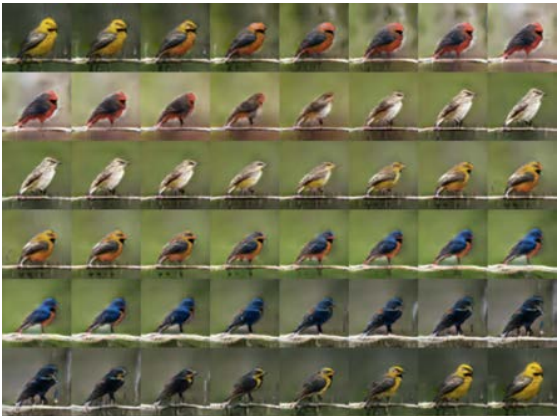

Figure 6: Figure showing interpolation between birds of six text descriptions by keeping the same posture.

**Fixed posture with varying details:** A bird with a particular posture and style could still have a lot of diversity in terms of fine-grained details and background. In this experiment, we show a way to sample them. This also shows the role and usefulness of condition augmentation in our model.

To keep the posture to be the same, as mentioned in the previous section, we sample a $z$ and generate $M_s$ which is held constant for this experiment. To vary the fine details, we consider a particular text description and pass it through $E$ and draw 5 samples from the Gaussian distribution around the text embedding applying the condition augmentation. Each of these 5 samples produce birds with the same posture (and style) but with varying amount of fine details and backgrounds as shown in Figure 5.

It can be observed from the second row of Figure 5 that for the red bird with black on its wing, even though all the birds have the same posture, no two birds are exactly the same. They all have varying amount of black on their wings and the length of the tail. Similar behavior can be seen in the fourth row, where all the birds are orange but with varying color saturation. This demonstrates that condition augmentation is positively adding to diversity by modeling the finer details of the birds in our model.

| Metric | GAN-INT-CLS | StackGAN-I | FusedGAN(ours) |
|---|---|---|---|
| Inception score | 2.88 ± .04 | 2.95 ± .02 | **3.00 ± .03** |
| Human rank | 1.60 | 1.91 | **3.12** |

Table 1: Inception scores and average human ranks

The GAN-INT-CLS does not have any additional control over sampling of text embedding. While the StackGAN shows that the condition augmentation helps in general in improving the diversity, it does not have a way to leverage it for controlled sampling. Using condition augmentation, our model can both improve the diversity and perform controlled sampling of birds with varying fine-grained details.

**Interpolation with the same posture but varying styles:** Our method also allows to interpolate between various styles by keeping the posture constant, as shown in Figure 6. To achieve this, we take two text-samples $t_1$ and $t_2$ and then pass them through $E$ to draw two samples from their respective Gaussian distributions. We obtain two samples each of $1 \times 128$ dimensions. Then, we interpolate between them to uniformly pick 8 samples, such that the first sample corresponds to $t_1$ and last one corresponds to $t_2$. We then draw a $z$ and generate a $M_s$ which is held constant for these 8 samples.

As described in our earlier sections and inference process, $M_s$ and the interpolated samples are given as inputs to $G_c$ to generate the conditional images. In Figure 6, we show some results of this interpolation experiment. The first and last image of each row correspond to two styles. All the images in between are interpolated. Moreover, the first image of each row is the same as the last image of the previous row. In this way, we interpolate between 5 different styles keeping the same posture. Note that the rows are interpolation between : $t_1 \rightarrow t_2$, $t_2 \rightarrow t_3$, $t_3 \rightarrow t_4$, $t_4 \rightarrow t_5$ and $t_5 \rightarrow t_1$ to complete the full cycle.

**Qualitative and quantitative comparison:** To quantitatively compare the results of our method with StackGAN stage-I and GAN-INT-CLS, we use the publicly available models from respective authors and compute the inception scores as shown in Table 1. We randomly sample 30k images for each model and compute the inception scores using the pre-trained model on CUB birds test set provided by StackGAN. Table 1 shows that our method obtains a slightly better inception score than the StackGAN, and beats GAN-INT-CLS with a significant margin. Since inception score has its own limitations in terms of fully evaluating fidelity and diversity, we also perform a user study to compare the results of our method with the two competing methods.

For this user study we randomly select 100 text descriptions and sample 8 images for every model. We show these images to 10 different people and ask them to score the fi-delity of the birds. None of the authors were part of the user study. The results of the user study shows that birds generated by our method have better visual quality compared to stage-I of StackGAN and GAN-INT-CLS. This can be partly attributed to the fact that our proposed model can leverage more training images with no paired text description, due to semi-supervised nature of our model. In addition to the limited visual results presented in Figure 4, we provide more visual results in supplementary material.

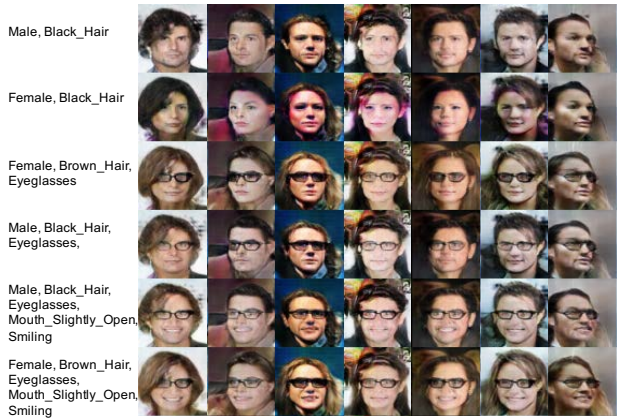

Figure 7: Generated visual examples illustrating the disentangling of style and structure in face synthesis.

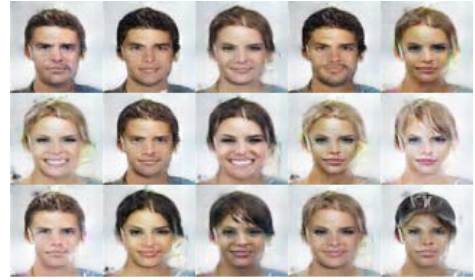

Figure 8: Generated visual example illustrating the synthesis of faces with the same structure but varying attributes.

## 5.2. Attribute-to-face generation

To further analyze the importance of disentangling and the structure prior, we evaluate its usefulness on attribute-to-face synthesis as shown in Figure 7. For this experiment, we use the celebA [7] dataset that has a 40 dimensional binary attribute vector annotated with each face image. We follow the same training protocol for building our proposed model, except that we do not augment the dataset with any more images without paired attributes. That is because celebA already has over 200k images, which are sufficient. We use the standard DCGAN architecture [9] for training and more details on this are provided in the supplementary material.

### 5.2.1 Results and Analysis

**Sampling with the same structure but varying attributes:** Similar to the experiment in birds generation, in this experiment, we draw a $z$ from $N(0, I)$ and keep the structure prior constant. We then give various attribute vectors as inputs to synthesize faces as shown in Figure 7. For every column in Figure 7, all the rows have the same pose and structure, but the synthesized faces vary as per the attributes. For every row, by default all the attributes are off and only the attributes that are shown next to each row are on.

We make several interesting observations from Figure 7. For example, in the first column, for the first two images, when the attribute for gender is switched from male to female, not only the pose but also some other characteristics of the face, such as hair style is also roughly preserved. Similarly for the third and fourth images in the same column, the faces look very similar. A closer inspection of these images reveal that there are subtle differences around the mouth and jaw bone areas, which distinguishes between male and female. More over, in the last column, it can be observed that the structure is preserved even for extreme poses. This further reaffirms that our model is able to successfully disentangle the style and structure.

Figure 8 presents a batch of faces generated from various random attributes but with the same structure prior. We can observe how different faces are generated with varying attributes, such as color and style of the hair, wearing a hat or not, skin tone, gender, etc., but they all look alike. Controlled sampling like this finds its use case in forensic science to synthesize similarly looking faces with varying attributes, which helps in making a portrait of a potential suspect.

## 6. Conclusion

We presented FusedGAN, a new deep generative model architecture for conditional image generation by fusing two generators, where one of them generates an unconditional image and the other generates a conditional image. The unconditional image generation can leverage additional training images without the corresponding conditions to learn a good structure prior. This in turn helps in synthesizing a better conditional image, as it takes the structure prior as part of its input in addition to the condition. The proposed model enjoys the strengths of stacking and disentangling without the need for separate training of stages or additional intermediate image supervision. Extensive analysis and experiments on text-to-image synthesis and attribute-to-face synthesis show that the model is able to successfully learn a disentangled representation for style and structure, and hence generate birds and faces with high fidelity, diversity, and more controllability in sampling.

## References

[1] B. Dai, D. Lin, R. Urtasun, and S. Fidler. Towards diverse and natural image descriptions via a conditional gan. *arXiv preprint arXiv:1703.06029*, 2017. 2

[2] I. Goodfellow, J. Pouget-Abadie, M. Mirza, B. Xu, D. Warde-Farley, S. Ozair, A. Courville, and Y. Bengio. Generative adversarial nets. In *Advances in neural information processing systems*, pages 2672–2680, 2014. 1

[3] K. Gregor, I. Danihelka, A. Graves, D. J. Rezende, and D. Wierstra. Draw: A recurrent neural network for image generation. *arXiv preprint arXiv:1502.04623*, 2015. 2

[4] I. Gulrajani, F. Ahmed, M. Arjovsky, V. Dumoulin, and A. Courville. Improved training of wasserstein gans. *arXiv preprint arXiv:1704.00028*, 2017. 1

[5] P. Isola, J.-Y. Zhu, T. Zhou, and A. A. Efros. Image-to-image translation with conditional adversarial networks. *arxiv*, 2016. 1

[6] D. P. Kingma and M. Welling. Auto-encoding variational bayes. *arXiv preprint arXiv:1312.6114*, 2013. 1

[7] Z. Liu, P. Luo, X. Wang, and X. Tang. Deep learning face attributes in the wild. In *Proceedings of International Conference on Computer Vision (ICCV)*, 2015. 8

[8] M. Mirza and S. Osindero. Conditional generative adversarial nets. *arXiv preprint arXiv:1411.1784*, 2014. 2

[9] A. Radford, L. Metz, and S. Chintala. Unsupervised representation learning with deep convolutional generative adversarial networks. *arXiv preprint arXiv:1511.06434*, 2015. 8

[10] S. Reed, Z. Akata, H. Lee, and B. Schiele. Learning deep representations of fine-grained visual descriptions. In *Proceedings of the IEEE Conference on Computer Vision and Pattern Recognition*, pages 49–58, 2016. 5

[11] S. Reed, Z. Akata, S. Mohan, S. Tenka, B. Schiele, and H. Lee. Learning what and where to draw. In *Advances in Neural Information Processing Systems*, 2016. 2

[12] S. Reed, Z. Akata, X. Yan, L. Logeswaran, B. Schiele, and H. Lee. Generative adversarial text to image synthesis. *arXiv preprint arXiv:1605.05396*, 2016. 2, 3, 5

[13] T. Salimans, I. Goodfellow, W. Zaremba, V. Cheung, A. Radford, and X. Chen. Improved techniques for training gans. In *Advances in Neural Information Processing Systems*, pages 2234–2242, 2016. 5

[14] G. Van Horn, S. Branson, R. Farrell, S. Haber, J. Barry, P. Ipeirotis, P. Perona, and S. Belongie. Building a bird recognition app and large scale dataset with citizen scientists: The fine print in fine-grained dataset collection. In *Proceedings of the IEEE Conference on Computer Vision and Pattern Recognition*, pages 595–604, 2015. 5

[15] C. Wah, S. Branson, P. Welinder, P. Perona, and S. Belongie. The Caltech-UCSD Birds-200-2011 Dataset. Technical report, 2011. 5

[16] X. Wang and A. Gupta. Generative image modeling using style and structure adversarial networks. In *ECCV*, 2016. 2, 3

[17] X. Yan, J. Yang, K. Sohn, and H. Lee. Attribute2image: Conditional image generation from visual attributes. corr abs/1512.00570, 2015. 2, 3

[18] H. Zhang, T. Xu, H. Li, S. Zhang, X. Huang, X. Wang, and D. Metaxas. Stackgan: Text to photo-realistic image synthesis with stacked generative adversarial networks. *arXiv preprint arXiv:1612.03242*, 2016. 2, 3, 5

[19] J.-Y. Zhu, T. Park, P. Isola, and A. A. Efros. Unpaired image-to-image translation using cycle-consistent adversarial networks. *arXiv preprint arXiv:1703.10593*, 2017. 1

[20] S. Zhu, S. Fidler, R. Urtasun, D. Lin, and C. C. Loy. Be your own prada: Fashion synthesis with structural coherence. 2, 3

# Semi-supervised FusedGAN for Conditional Image Generation
## Supplementary Material

Navaneeth Bodla[1]    Gang Hua[2]    Rama Chellappa[1]
Center For Automation Research, University of Maryland, College Park [1]   Microsoft Research[2]
{nbodla,rama}@umiacs.umd.edu   ganghua@microsoft.com

## Abstract

*In this supplementary material, we present: 1) additional visual results for text-to-image synthesis for birds and 2) architecture details of our model for training attribute-to-face synthesis. We also attach a short demo video showcasing the various controlled sampling examples of our model.*

## 1. Attribute-to-face synthesis: Architecture

As mentioned in section 5.2 of the main paper, our architecture is inspired from DCGAN. The generator $G_s$ takes uniform random noise from $[-1, 1]$ of size $1 \times 128$ as input to first produce a tensor of $8 \times 8 \times 512$. To do this, a fully connected layer with batch normalization, relu and reshape is used to produce a tensor of $4 \times 4 \times 1024$. Next, transposed convolution is performed with a $5 \times 5$ filter and stride 2. Finally batch normalization and relu are used to produce the output.

$G_u$ takes the tensor of $8 \times 8 \times 512$ from $G_s$ as input to produce a final image of size $64 \times 64 \times 3$. This is achieved by a series of 3 transposed convolution operations which successively produce tensors of shape : $16 \times 16 \times 256$, $32 \times 32 \times 128$ and $64 \times 64 \times 3$. The transposed convolution operations are performed with a $5 \times 5$ filter, stride 2. The first two transposed convolutions are followed by batch normalization and relu and for the final image generation, tanh is used without any batch normalization.

$G_c$ takes the tensor of $8 \times 8 \times 512$ from $G_s$ and 40 dimensional attribute vector as inputs to produce a final conditional image of size $64 \times 64 \times 3$. For this, the attribute vector is passed through fully connected layers to produce 40 dimensional $\mu$ and $\sigma$ for the Gaussian distribution centered around the input attribute vector. Next, a sample is drawn from this distribution and spatially repeated to produce a tensor of $8 \times 8 \times 40$. This tensor is then concatenated with the output of $G_s$ to produce a tensor of $8 \times 8 \times 552$. Finally the operations mentioned above in $G_u$ are performed to produce the conditional image.

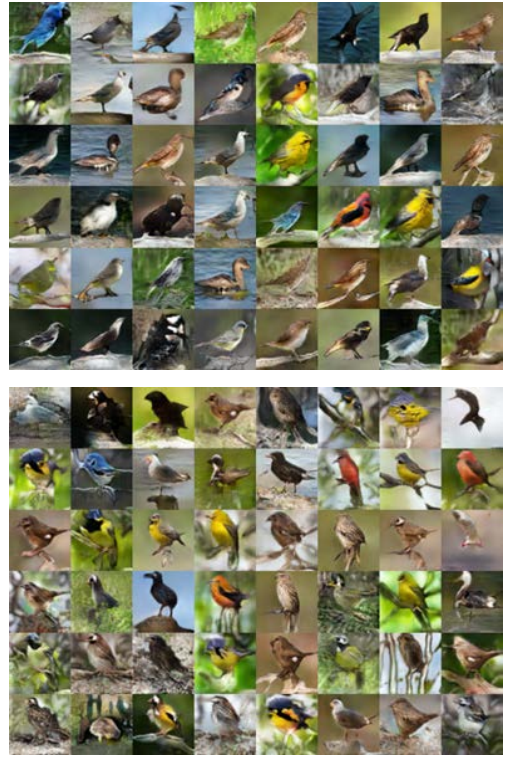

Figure 1: Random batches of samples with same posture but varying styles

Architectures for $D_c$ and $D_u$ are same as the discriminator of DCGAN, except that for $D_c$ we spatially repeat the condition and concatenate with the $8 \times 8$ spatial resolution feature map in the discriminator.

## 2. Text-to-image Synthesis

We show extensive results of our model in Figures 1 and 2 - 4. Figure 1 shows two random batches of synthesized birds with same posture. In Figures 2 - 3 we draw 48 samples per text description with varying posture and visualize the synthesized birds.

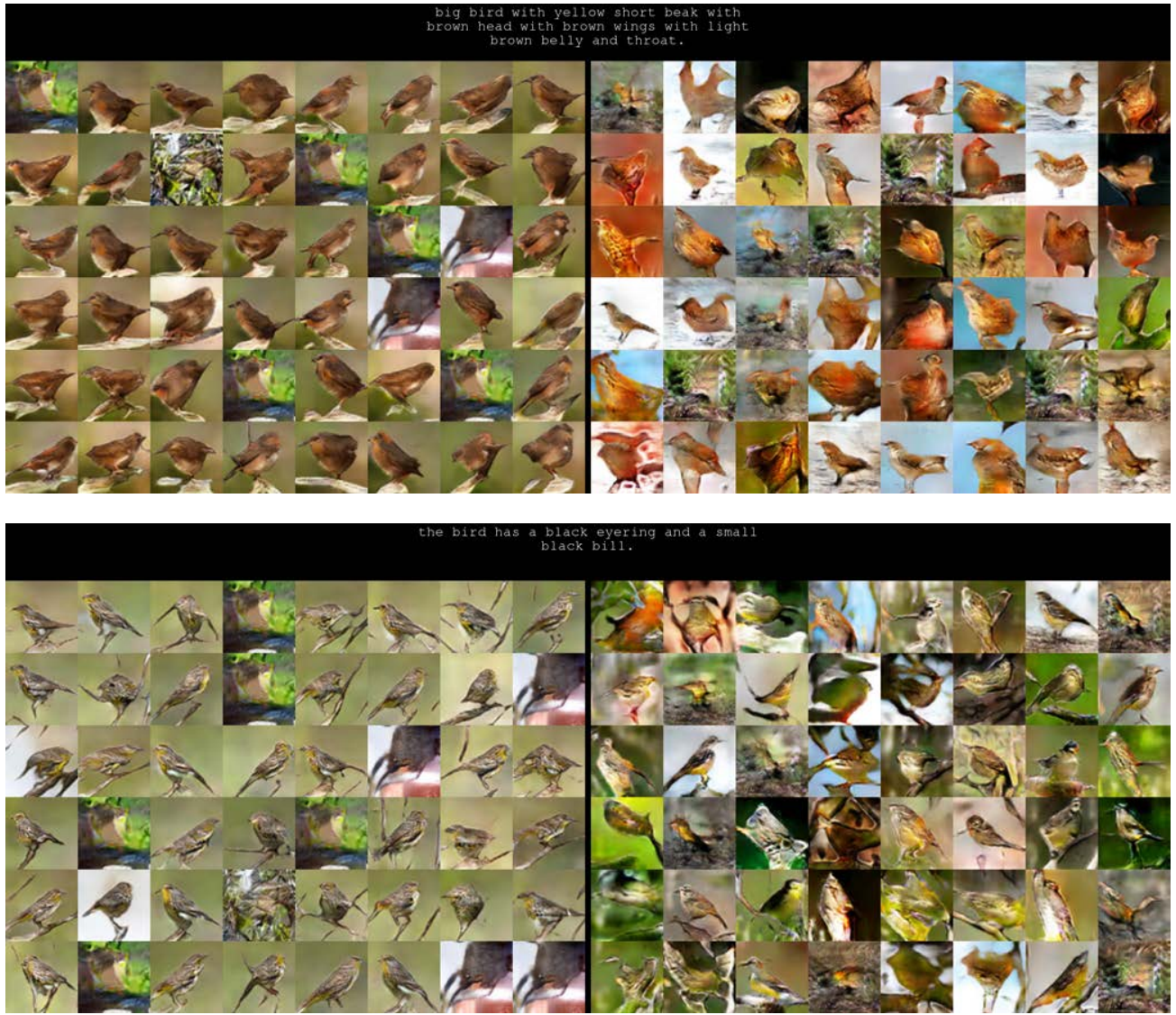

Figure 2: Visual comparison. Images on the left are from our FusedGAN and right are from StackGAN stage I

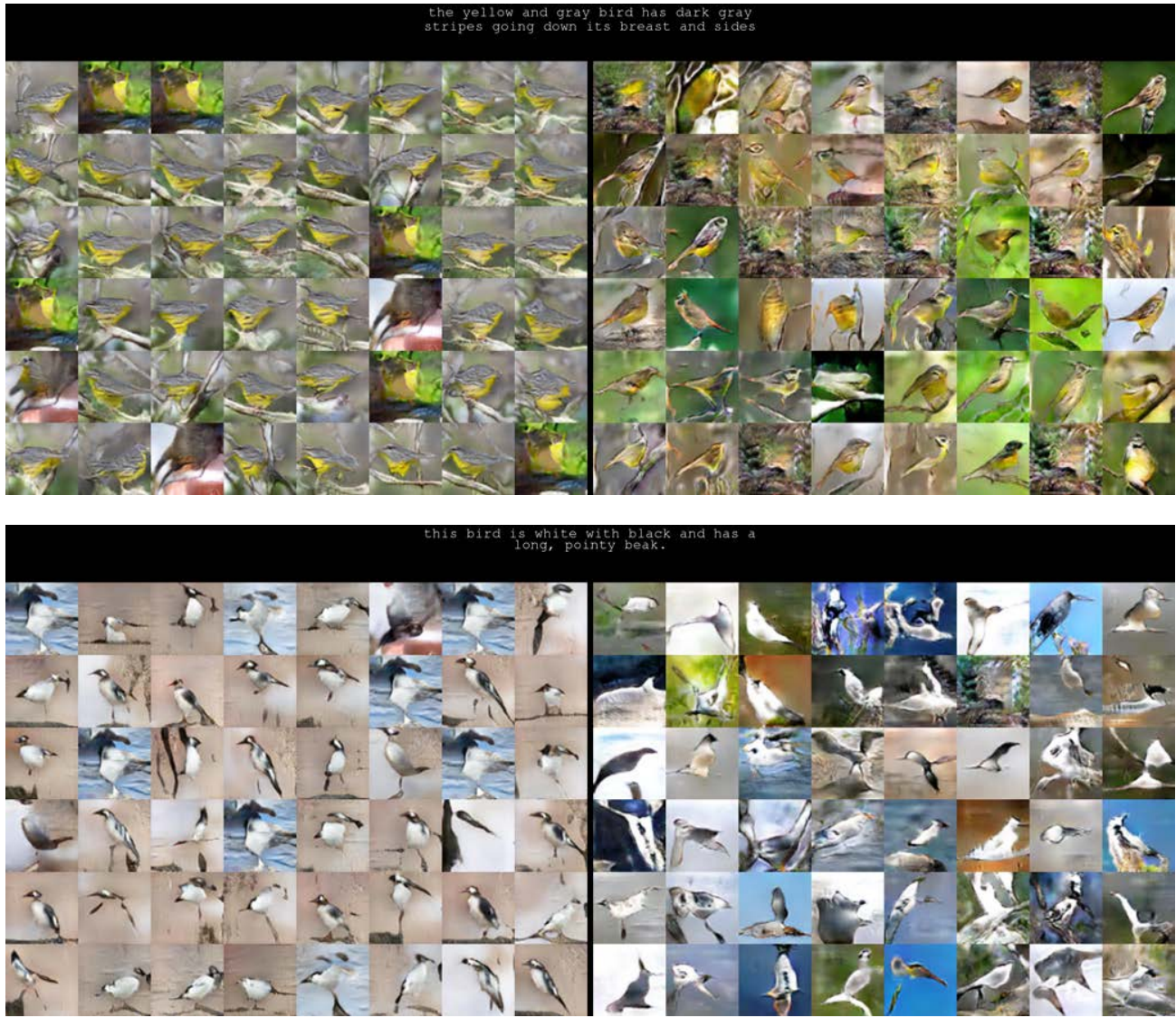

Figure 3: Visual comparison. Images on the left are from our FusedGAN and right are from StackGAN stage I

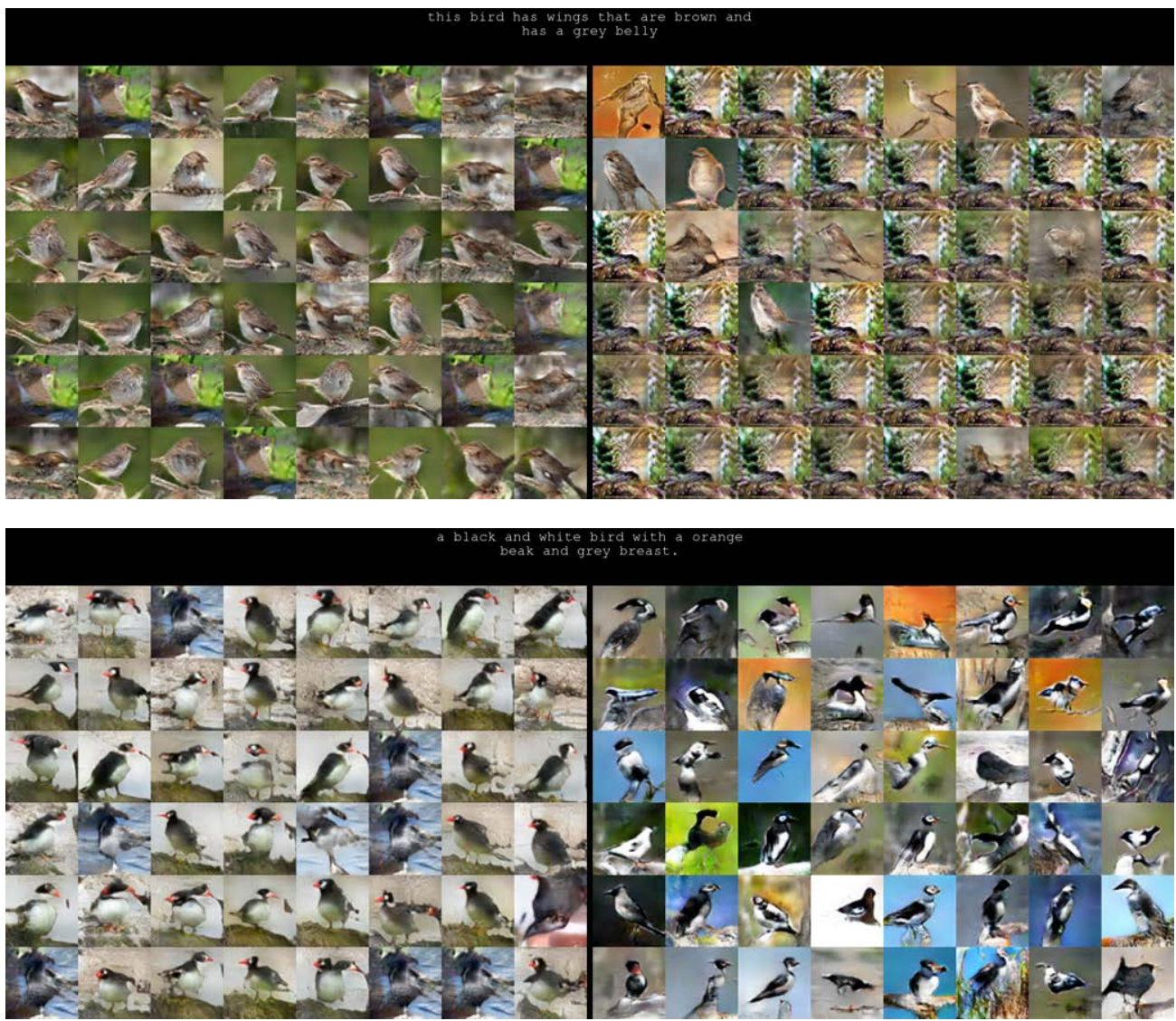

Figure 4: Visual comparison. Images on the left are from our FusedGAN and right are from StackGAN stage I